\documentclass[11pt,a4paper]{article}
\usepackage[hyperref]{emnlp2018}
\usepackage{times}
\usepackage{latexsym}
\usepackage{subfigure}
\usepackage{graphicx}
\usepackage{multirow}
\usepackage{booktabs}
\usepackage{comment}
\usepackage{algpseudocode,algorithm,algorithmicx}
\usepackage{amsfonts}
\usepackage{url}
\usepackage{amsmath}
\newcommand{\norm}[1]{\left\lVert#1\right\rVert}

\DeclareMathOperator*{\argmin}{arg\,min}

\aclfinalcopy % Uncomment this line for the final submission
\def\aclpaperid{1234} %  Enter the acl Paper ID here

\title{One-Shot Relational Learning for Knowledge Graphs}

\author{
 Wenhan Xiong$^\dagger$,
 Mo Yu$^\ast$, 
 Shiyu Chang$^\ast$, 
 Xiaoxiao Guo$^\ast$, 
 William Yang Wang$^\dagger$
\\ 
 $^\dagger$ University of California, Santa Barbara\\
 $^\ast$ IBM Research\\
 \{xwhan, william\}@cs.ucsb.edu, yum@us.ibm.com, \{shiyu.chang, xiaoxiao.guo\}@ibm.com  
 }

\date{}

\begin{document}
\maketitle
\begin{abstract}
Knowledge graphs (KGs) are the key components of various natural language processing applications. To further expand KGs' coverage, previous studies on knowledge graph completion usually require a large number of training instances for each relation. However, we observe that long-tail relations are actually more common in KGs and those newly added relations often do not have many known triples for training. In this work, we aim at predicting new facts under a challenging setting where only one training instance is available. We propose a one-shot relational learning framework, which utilizes the knowledge extracted by embedding models and learns a matching metric by considering both the learned embeddings and one-hop graph structures.  Empirically, our model yields considerable performance improvements over existing embedding models, and also eliminates the need of re-training the embedding models when dealing with newly added relations.\footnote{Code and datasets could be found at \url{https://github.com/xwhan/One-shot-Relational-Learning}.}

%   We also introduce two newly constructed datasets for one-shot learning on KGs.\william{The abstract is very head-heavy at this time. It talks too much about the background and other people's work. You might need to readjust accordingly.}
\end{abstract}

%%%%%%%%%%%%%%%%%%%%%%%%%%%%%%%%%%%%%%%%%%%%%%%%%%%%%%%%%%%%%%%%%%%%%%%%%%%%%%%%%%%%%%%%%%%%%%%%%%
\section{Introduction}
Large-scale knowledge graphs~\cite{suchanek2007yago,vrandevcic2014wikidata,bollacker2008freebase,auer2007dbpedia,carlson2010toward} represent every piece of information as binary relationships between entities, usually in the form of triples \emph{i.e.} \textit{(subject, predicate, object)}. This kind of structured knowledge is essential for many downstream applications such as Question Answering and Semantic Web. 

\begin{figure}[t]
\centering
\includegraphics[width=1.0\linewidth]{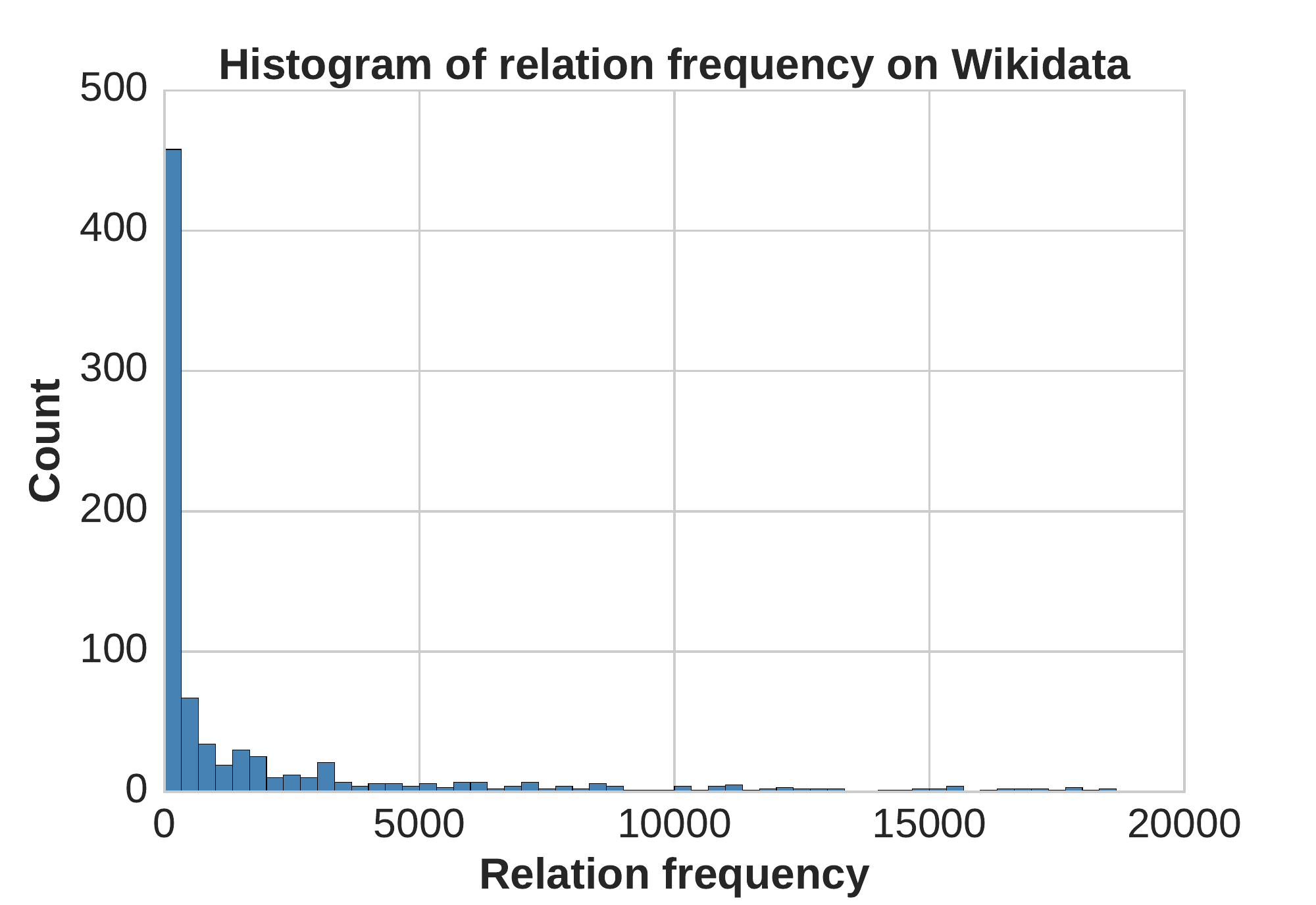}
\caption{The histogram of relation frequencies in Wikidata. There are a large portion of relations that only have a few triples.}
\label{dist}
\vspace{-0.1in}
\end{figure}

Despite KGs' large scale, they are known to be highly incomplete~\cite{min2013distant}.  To automatically complete KGs, extensive research efforts ~\cite{nickel2011three, bordes2013translating,yang2014embedding,trouillon2016complex,lao2010relational,neelakantan2015compositional,xiong2017deeppath,das2017go,chen2018variational} have been made to build relational learning models that could infer missing triples by learning from existing ones.  These methods explore the statistical information of triples or path patterns to infer new facts of existing relations; and have achieved considerable performance on various public datasets.

% \textcolor{red}{STORY: to achieve KBC on real-world scenario, we need to handle with two problems: (1) sparsity; (2) new added relations (dynamic property). Talk about each problem, its importance and why previous work cannot handle.}

However, those datasets (\emph{e.g.} FB15k, WN18) used by previous models mostly only cover common relations in KGs. For more practical scenarios, we believe the desired KG completion models should handle two key properties of KGs. First, as shown in Figure~\ref{dist}, a large portion of KG relations are actually long-tail.  In other words, they have very few instances. But intuitively, the fewer training triples that one relation has, the more KG completion techniques could be of use. Therefore, it is crucial for models to be able to complete relations with limited numbers of triples.  However, existing research usually assumes the availability of sufficient training triples for all relations, which limits their usefulness on sparse long-tail relations.

Second, to capture up-to-date knowledge, real-world KGs are often dynamic and evolving at any given moment. New relations will be added whenever new knowledge is acquired.  If a model can predict new triples given only a small number of examples, a large amount of human effort could be spared.  However, to predict target relations, previous methods usually rely on well-learned representations of these relations.  In the dynamic scenario, the representations of new relations cannot be sufficiently trained given limited training instances, thus the ability to adapt to new relations is also limited for current models.

In contrast to previous methods, we propose a model that depends only on the entity embeddings and local graph structures. Our model aims at learning a matching metric that can be used to discover more similar triples given one reference triple. The learnable metric model is based on a permutation-invariant network that effectively encodes the one-hop neighbors of entities, and also a recurrent neural network that allows multi-step matching. Once trained, the model will be able to make predictions about any relation while existing methods usually require fine-tuning to adapt to new relations. With two newly constructed datasets, we show that our model can achieve consistent improvement over various embedding models on the one-shot link prediction task. 

In summary, our contributions are three-fold:
\begin{itemize}
    \item We are the first to consider the long-tail relations in the link prediction task and formulate the problem as few-shot relational learning;
    \item We propose an effective one-shot learning framework for relational data, which achieves better performance than various embedding-based methods;
    \item We also present two newly constructed datasets for the task of one-shot knowledge graph completion.
\end{itemize}

\section{Related Work}
\paragraph{Embedding Models for Relational Learning}
Various models have been developed to model relational KGs in continous vector space and to automatically infer missing links. RESCAL~\cite{nickel2011three} is one of the earlier work that models the relationship using tensor operations. ~\newcite{bordes2013translating} proposed to model relationships in the 1-D vector space. Following this line of research, more advanced models such as DistMult~\cite{yang2014embedding}, ComplEx~\cite{trouillon2016complex} and ConvE~\cite{dettmers2017convolutional} have been proposed. These embedding-based models usually assume enough training instances for all relations and entities and do not pay attention to those sparse symbols. More recently, several models~\cite{shi2017open,xie2016representation} have been proposed to handle unseen entities by leveraging text descriptions. In contrast to these approaches, our model deals with long-tail or newly added relations and focuses on one-shot relational learning without any external information, such as text descriptions of entities or relations.

\paragraph{Few-Shot Learning}
Recent deep learning based few-shot learning approaches fall into two main categories: (1) \emph{metric based approaches} \cite{koch2015siamese,vinyals2016matching,snell2017prototypical,yu2018diverse}, which try to learn 
generalizable metrics and the corresponding matching functions from a set of training tasks.
Most methods in this class adopt the general matching framework proposed in deep siamese network~\citep{koch2015siamese}. One example is the Matching Networks~\citep{vinyals2016matching}, which make predictions by comparing the input example with a small labeled support set;
(2) \emph{meta-learner based approaches}~\cite{ravi2017optimization,munkhdalai2017meta,finn2017model,li2017meta}, which aim to learn the optimization of model parameters (by either outputting the parameter updates or directly predicting the model parameters) given the gradients on few-shot examples.
One example is the LSTM-based meta-learner~\citep{ravi2017optimization}, which learns the step size for each dimension of the stochastic gradients. 
Besides the above categories, there are also some other styles of few-shot learning algorithms, \emph{e.g.} Bayesian Program Induction~\citep{lake2015human}, which represents concepts as simple programs that best explain observed examples under a Bayesian criterion.  

Previous few-shot learning research mainly focuses on vision and imitation learning~\cite{duan2017one} domains. In the language domain, \newcite{yu2018diverse} proposed a multi-metric based approach for text classification.
To the best of our knowledge, this work is the first research on few-shot learning for knowledge graphs.

\section{Background}

\subsection{Problem Formulation}
Knowledge graphs $\mathcal{G}$ are represented as a collection of triples $\{(h,r,t)\} \subseteq \mathcal{E} \times \mathcal{R} \times \mathcal{E}$, where $\mathcal{E}$ and $\mathcal{R}$ are the entity set and relation set. The task of knowledge graph completion is to either predict unseen relations $r$ between two existing entities: $(h,?,t)$ or predict the tail entity $t$ given the head entity and the query relation: $(h,r,?)$. As our purpose is to infer unseen facts for newly added or existing long-tail relations, we focus on the latter case. In contrast to previous work that usually assumes enough triples for the query relation are available for training, this work studies the case where only one training triple is available. To be more specific, the goal is to rank the true tail entity $t_{true}$ higher than other candidate entities $t \in \mathcal{C}_{h,r}$, given only an example triple $(h_0,r,t_0)$. The candidates set is constructed using the entity type constraint~\cite{toutanova2015representing}. It is also worth noting that when we predict new facts of the relation $r$, we only consider a closed set of entities, \emph{i.e.} no unseen entities during testing. For open-world settings where new entities might appear during testing, external information such as text descriptions about these entities are usually required and we leave this to future work.

% It is worth noting that once a new triple with new relation is added, new entity may also be introduced. It is also the case when some entity only co-occur with some particular sparse relation. This is a more challenging setting and we leave it to future work. 

\subsection{One-Shot Learning Settings}
This section describes the settings for the training and evaluation of our one-shot learning model.

The goal of our work is to learn a metric that could be used to predict new facts with one-shot examples. Following the standard one-shot learning settings~\cite{vinyals2016matching,ravi2017optimization}, we assume access to a set of training tasks. In our problem, each training task corresponds to a KG relations $r\in \mathcal{R}$, and has its own training/testing triples: $T_r=\{D_r^{train},D_r^{test}\}$.
This task set is often denoted as the \emph{meta-training} set, $\mathbb{T}_{meta-train}$. 

To imitate the one-shot prediction at evaluation time, there is only one triple $(h_0, r, t_0)$ in each $D_r^{train}$.
The $D_r^{test}=\{(h_i,r,t_i,\mathcal{C}_{h_i,r})\}$ consists of the testing triples of $r$ with ground-truth tail entities $t_i$ for each query $(h_i, r)$, and the corresponding tail entity candidates $\mathcal{C}_{h_i,r}=\{t_{ij}\}$ where each $t_{ij}$ is an entity in $\mathcal{G}$.
The metric model can thus be tested on this set by ranking the candidate set $\mathcal{C}_{h_i,r}$ given the test query $(h_i, r)$ and the labeled triple in  $D_r^{train}$.
We denote an arbitrary ranking-loss function as $\ell_{\theta}(h_i,r,t_i|\mathcal{C}_{h_i,r}, D_r^{train})$, 
where $\theta$ represents the parameters of our metric model.
This loss function indicates how well the metric model  works on tuple $(h_i,r,t_i,\mathcal{C}_{h_i,r})$ while observing only one-shot data from $D_r^{train}$.
The objective of training the metric model, \emph{i.e.} the meta-training objective, thus becomes:
\begin{equation}
\small
    \min_{\theta}\mathbb{E}_{T_r}\left [ \sum_{(h_i,r,t_i,\mathcal{C}_{h_i,r}) \in D_r^{test}}\frac{\ell_{\theta}(h_i,r,t_i|\mathcal{C}_{h_i,r}, D_r^{train})}{\vert D_r^{test} \vert}\right ],
    % \theta = \argmin_{\theta}\mathbb{E}_{T_r \sim \mathbb{T}_{meta-train}}\sum_{(h_i,r,t_i,\mathcal{C}_{h_i,r}) \in D_r^{test}}\ell_{\theta}(h_i,r,t_i|\mathcal{C}_{h_i,r}, D_r^{train}),
\end{equation}
where $T_r$ is sampled from the meta-training set $\mathbb{T}_{meta-train}$, and $\vert D_r^{test} \vert$ denotes the number of tuples in $D_r^{test}$.

Once trained, we can use the model to make predictions on new relations $r' \in \mathcal{R'}$, which is called the \emph{meta-testing} step in literature. These meta-testing relations are unseen from meta-training, \emph{i.e.} $\mathcal{R'} \cap \mathcal{R} = \phi$.
Each meta-testing relation $r'$ also has its own one-shot training data $D_{r'}^{train}$ and testing data $D_{r'}^{test}$, defined in the same way as in meta-training.
These meta-testing relations form a meta-test set $\mathbb{T}_{meta-test}$.

Moreover, we leave out a subset of relations in $\mathbb{T}_{meta-train}$ as the meta-validation set $\mathbb{T}_{meta-validation}$.
Because of the assumption of one-shot learning, the meta-testing relations do not have validation sets like in the traditional machine learning setting. Otherwise, the metric model will actually see more than one-shot labeled data during meta-testing, thus the one-shot assumption is violated.

Finally, we assume that the method has access to a \emph{background knowledge graph} $\mathcal{G}'$, which is a subset of $\mathcal{G}$ with all the relations from $\mathbb{T}_{meta-train}$, $\mathbb{T}_{meta-validation}$ and $\mathbb{T}_{meta-test}$ removed.

\section{Model}

\begin{figure*}[t]
\centering
\includegraphics[width=1.02\linewidth]{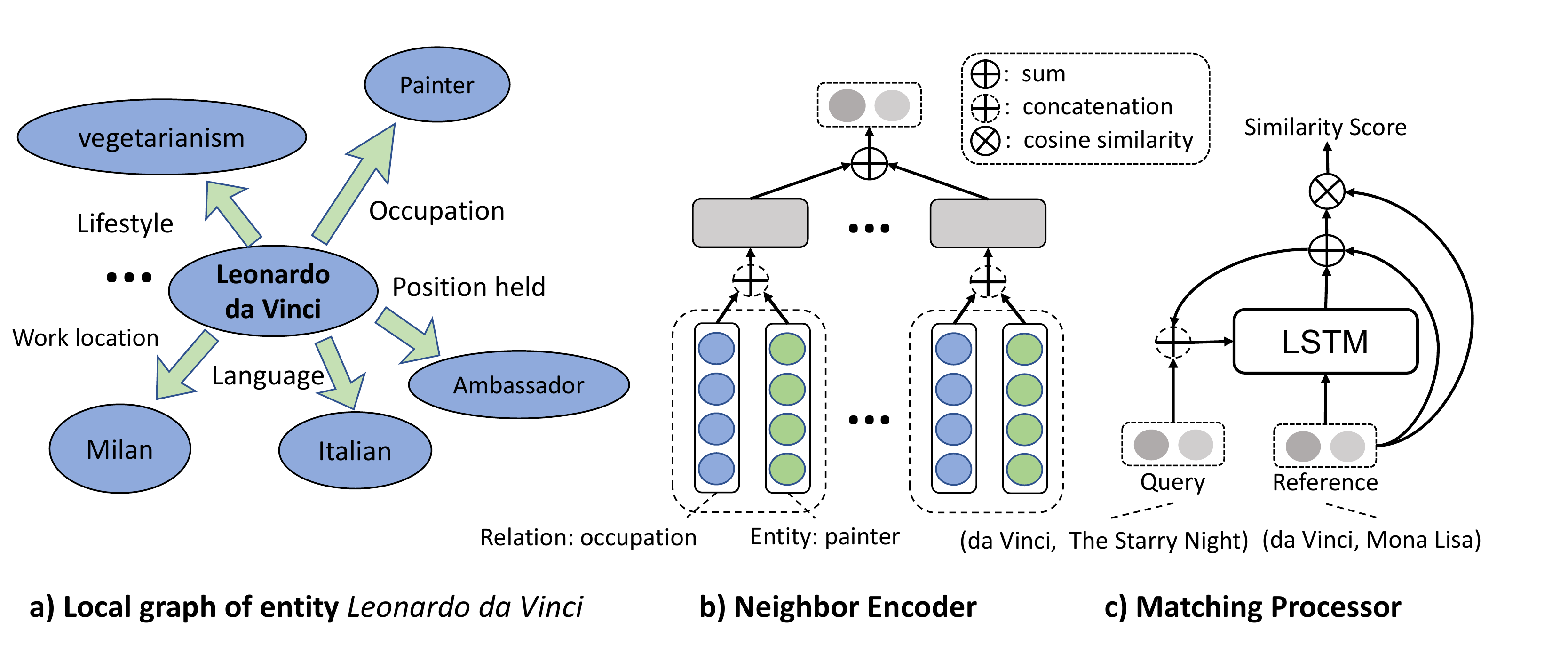}
\vspace{-0.5in}
\caption{a) and b): Our \textbf{neighbor encoder} operating on entity \textit{Leonardo da Vinci}; c): The \textbf{matching processor}.}
\label{neighbor}
\end{figure*}

In this section, we describe the proposed model for similarity metric learning and also the corresponding loss function $\ell$ we use to train our model. 

The core of our proposed model is a similarity function $\mathcal{M}((h,t),(h',t')|\mathcal{G}')$. 
Thus for any query relation $r$, as long as there is one known fact
$(h_0, r, t_0)$, the model could predict the likelihood of testing triples $\{(h_i,r,t_{ij})|t_{ij} \in \mathcal{C}_{h_i,r}\}$, based on the matching score between each $(h_i,t_{ij})$ and $(h_0,t_0)$.
% The inputs of our metric model will be the reference entity pair $(h_0,t_0)$, a query entity pair $(h_i,t_{ij})$.
The implementation of the above matching function involves two sub-problems: (1) the representations of entity pairs; and (2) the comparison function between two entity-pair representations. Our overall model, as shown in Figure~\ref{neighbor}, deals with the above two problems with two major components respectively:\\
$\bullet$ \textbf{Neighbor encoder} (Figure~\ref{neighbor}b), aims at utilizing the local graph structure to better represent entities. In this way, the model can leverage more information that KG provides for every entity within an entity pair.\\
$\bullet$ \textbf{Matching processor} (Figure~\ref{neighbor}c), takes the vector representations of any two entity pairs from the neighbor encoder; then performs multi-step matching between two entity-pairs and outputs a scalar as the similarity score.

% The overall model, as shown in Figure~\ref{neighbor}, includes two major components. In order to leverage more information that KG provides for every entity, the first component, which we refer as the \textbf{neighbor encoder} (Figure~\ref{neighbor}b), aims to encode every entity's local connections with other entities. By applying this encoder to every entity within an entity pair, we can get a concatenated vector that encodes the entity pair's local graph structures. With the vector representations of any two entity pairs, we then use the second component, the \textbf{matching processor} (Figure~\ref{neighbor}c), to perform multi-step matching and output a scalar as the similarity score. More details of these two components are described separately in the following paragraphs.

\subsection{Neighbor Encoder} 
This module is designed to enhance the representation of each entity with its local connections in knowledge graph.

Although the entity embeddings from KG embedding models \cite{bordes2013translating,yang2014embedding} already have relational information encoded, previous work~\cite{neelakantan2015compositional,lin2015modeling,xiong2017deeppath} showed that explicitly modeling the structural patterns, such as paths, is usually beneficial for relationship prediction. 
In view of this, we propose to use a neighbor encoder to incorporate graph structures into our metric-learning model. 
In order to benefit from the structural information while maintaining the efficiency to easily scale to real-world large-scale KGs, our neighbor encoder only considers entities' local connections, \emph{i.e.} the one-hop neighbors.

% For any query relation $r$, we consider one known triple $(h_0,r,t_0)$ and any other test triple from $\{(h_i,r,t_{ij})|t_{ij} \in \mathcal{C}_{h_i,r}\}$. The inputs of our metric model will be the reference entity pair $(h_0,t_0)$, a query entity pair $(h_i,t_{ij})$ and the knowledge graph $\mathcal{G}$.  
% To match these two entity-pairs $(h_0,t_0)$ and $(h_i,t_{ij}), t_{ij} \in \mathcal{C}_{h_i,r}$, a simple solution could be metric models that only consider the entity embeddings learned by a KB embedding model, as these embedding models are actually designed to model relationships. However, as shown by previous work~\cite{neelakantan2015compositional,xiong2017deeppath}, explicitly modeling the graph patterns such as paths is usually beneficial. In view of this, we propose this neighbor encoder to encode graph structures into our model. But instead of modeling the paths, which can be expensive to extract from large-scale KGs, we only consider entities' local connections, \emph{i.e.} one-hop neighbors. This enables our model to easily scale to real-world KGs while at the same time allows us to achieve better performance than simple embedding-based matching.

For any given entity $e$, its local connections form a set of (\textit{relation, entity}) tuples. As shown in Figure~\ref{neighbor}a, for the entity \textit{Leonardo da Vinci}, one of such tuples is (\textit{occupation, painter}). We refer this neighbor set as as $\mathcal{N}_{e}=\{(r_k,e_k)|(e,r_k,e_k)\in \mathcal{G}'\}$. 
The purpose of our neighbor encoder is to encode $\mathcal{N}_e$ and output a vector as the latent representation of $e$. Because this is a problem of encoding sets with varying sizes, we hope the encoding function can be (1) invariant to permutations and also (2) insensitive to the size of the neighbor set.
% To effectively encode all the neighbors, the desired encoder should satisfy two important requirements. First, the output should be invariant to set permutations. Second, the scale of the output vector should not be affected by the cardinality of the neighbor set. 
Inspired by the results from \cite{zaheer2017deep}, we use the following function $f$ that satisfies the above properties:
\begin{equation}
    f(\mathcal{N}_e) = \sigma(\frac{1}{|\mathcal{N}_e|}\sum_{(r_k,e_k)\in \mathcal{N}_e}C_{r_k,e_k}).
\end{equation}
where $C_{r_k,e_k}$ is the feature representation of a relation-entity pair $(r_k, e_k)$ and $\sigma$ is the activation function. In this paper we set $\sigma=\mathrm{tanh}$ which achieves the best performance on $\mathbb{T}_{meta-validation}$.

To encode every tuple $(r_k,e_k)\in \mathcal{N}_e$ into $C_{r_k,e_k}$, we first use an embedding layer $\mathbf{emb}$ with dimension $d$ (which can be pre-trained using existing embedding-based models) to get the vector representations of $r_k$ and $e_k$:
\begin{align}
    v_{r_k} = \mathbf{emb}(r_k), v_{e_k} = \mathbf{emb}(e_k) \nonumber
    % [v_{r_k}, v_{e_k}] = \mathrm{Dropout}(\mathrm{Embedding}([r_k,e_k])).
\end{align}
Dropout~\cite{srivastava2014dropout} is applied here to the vectors $v_{r_k}, v_{e_k}$ to achieve better generalization. We then apply a feed-forward layer to encode the interaction within this tuple:
\begin{equation}
    C_{r_k,e_k} = W_c (v_{r_k} \oplus v_{e_k}) + b_c,
\end{equation}
where $W_c \in R^{d \times 2d}, b_c \in R^{d}$ are parameters to be learned and $\oplus$ denotes concatenation.

\begin{figure}[t]
\centering
\includegraphics[width=1.0\linewidth]{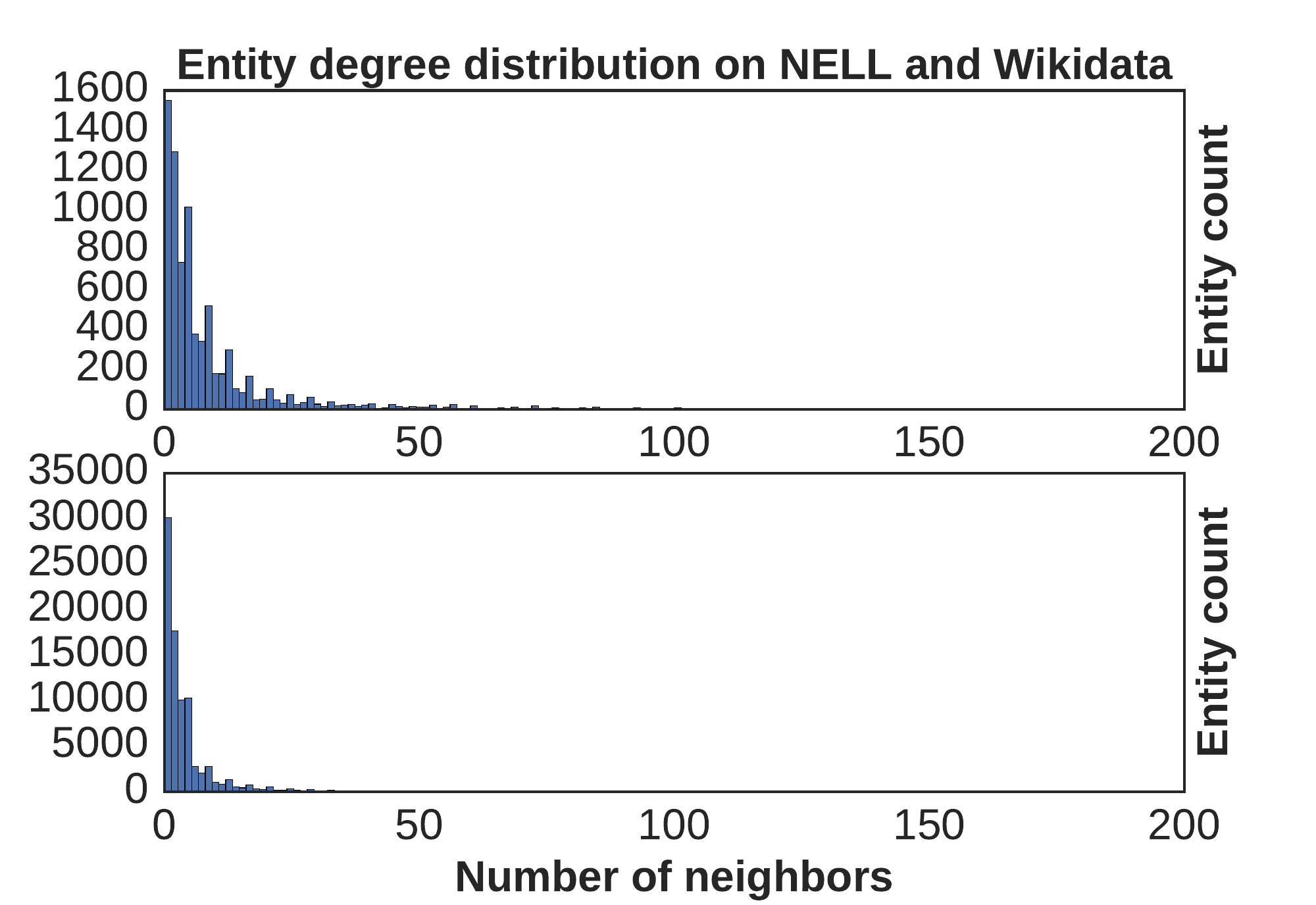}
\vspace{-1ex} 
\caption{The distribution of entities' degrees (numbers of neighbors) on our two datasets. 
Since we work on closed-set of entities, we draw the figure by considering the intersection between entities in our background knowledge $\mathcal{G}'$ and the entities appearing in $\mathbb{T}_{meta-train}$, $\mathbb{T}_{meta-validation}$ or $\mathbb{T}_{meta-test}$. Note that all triples in $\mathbb{T}_{meta-train}$, $\mathbb{T}_{meta-validation}$ or $\mathbb{T}_{meta-test}$ are removed from $\mathcal{G}'$. Upper: NELL; Lower: Wikidata.}
\label{degree}
\end{figure}

To enable batching during training, we manually specify the maximum number of neighbors and use all-zero vectors as ``dummy'' neighbors. Although different entities have different degrees (number of neighbors), the degree distribution is usually very concentrated, as shown in Figure~\ref{degree}. We can easily find a proper bound as the maximum number of neighbors to batch groups of entities.

% Although different entities usually have different numbers of neighbors(different degrees), the distribution 

The neighbor encoder module we propose here is similar to the Relational Graph Convolutional Networks~\cite{schlichtkrull2017modeling} in the sense that we also use the shared kernel $\{W_c,b_c\}$ to encode the neighbors of different entities. But unlike their model that operates on the whole graph and performs multiple steps of information propagation, we only encode the local graphs of the entities and perform one-step propagation. This enables us to easily apply our model to large-scale KGs such as Wikidata. Besides, their model also does not operate on pre-trained graph embeddings.
We leave the investigation of other graph encoding strategies, e.g. \cite{xu2018graph2seq,song2018graph}, to future work.

\subsection{Matching Processor} Given the neighbor encoder module, now we discuss how we can do effective similarity matching based on our recurrent matching processor. By applying $f(\mathcal{N}_e)$ to the reference entity pair $ (h_0,t_0)$ and any query entity pair $(h_i,t_{ij})$, we get two neighbor vectors for each: $[f(\mathcal{N}_{h_0});f(\mathcal{N}_{t_0})]$ and $[f(\mathcal{N}_{h_i});f(\mathcal{N}_{t_{ij}})]$. To get a similarity score that can be used to rank $(h_i,t_{ij})$ among other candidates, we can simply concatenate the $f(\mathcal{N}_{h})$ and $f(\mathcal{N}_{t})$ in each pair to form a single pair representation vector, and calculate the cosine similarity between pairs. However, this simple metric model turns out to be too shallow and does not give good performance. To enlarge our model's capacity, we leverage a LSTM-based~\cite{hochreiter1997long} recurrent ``processing'' block~\cite{vinyals2015order,vinyals2016matching} to perform multi-step matching. Every process step is defined as follows:
\begin{align}
   h_{k+1}^{'}, c_{k+1} &= LSTM(q, [h_k \oplus s,c_k]) \nonumber \\
   h_{k+1} &= h_{k+1}^{'} + q \nonumber \\
   score_{k+1} &= \frac{h_{k+1} \odot s }{\norm{h_{k+1}}\norm{s}},\label{eq:scoring}
\end{align}
where $LSTM(x,[h,c])$ is a standard LSTM cell with input $x$, hidden state $h$ and cell state $c$, and $s = f(\mathcal{N}_{h_0})\oplus f(\mathcal{N}_{t_0}), q = f(\mathcal{N}_{h_i}) \oplus f(\mathcal{N}_{t_{ij}})$ are the concatenated neighbor vectors of the reference pair and query pair. After $K$ processing steps\footnote{$K$ is a hyperparameter to be tuned.}, we use $score_K$ as the final similarity score between the query and support entity pair. For every query $(h_i,r,?)$, by comparing $(h_i,t_{ij})$ with $(h_0,t_0)$, we can get the ranking scores for every $t_{ij} \in \mathcal{C}_{h_i,r}$.

\begin{algorithm}[t] 
\small
    \caption{One-shot Training}
    \label{alg:training}
    \begin{algorithmic}[1]
        \State 
        \textbf{Input:}\\ 
        a) Meta-training task set $\mathbb{T}_{meta-training}$; \\
        b) Pre-trained KG embeddings (excluding relation in $\mathbb{T}_{meta-training}$);\\
        c) Initial parameters $\theta$ of the metric model;
        
        \For{epoch = 0:M-1}
            \State Shuffle the tasks in $\mathcal{T}_{meta-learning}$
            \For{$T_r$ in $\mathcal{T}_{meta-learning}$}
                \State Sample one triple as the reference
                \State Sample a batch $B^{+}$ of query triples 
                \State Pollute the tail entity of query triples to get $B^{-}$
                \State Calculate the matching scores for triple in $B^{+}$ and $B^{-}$
                \State Calculate the batch loss $\mathcal{L} = \sum_B \ell$
                \State Update $\theta$ using gradient $g \propto \nabla \mathcal{L}$
            \EndFor
        \EndFor
        
    \end{algorithmic}
\end{algorithm}
\subsection{Loss Function and Training} For a query relation $r$ and its reference/training triple $(h_0,r,t_0)$, we collect a group of positive (true) query triples $\{(h_i,r,t_i^{+})|(h_i,r,t_i^{+})\in \mathcal{G}\}$ and construct another group negative (false) query triples $\{(h_i,r,t_i^{-})|(h_i,r,t_i^{-})\not\in \mathcal{G}\}$ by polluting the tail entities. Following previous embedding-based models, we use a hinge loss function to optimize our model:
\begin{equation}
    \ell_{\theta}
    = max(0, \gamma + score^{-}_\theta - score^{+}_\theta),
\end{equation}
where $score^{+}_\theta$ and $score^{-}_\theta$ are scalars calculated by comparing the query triple $(h_i,r,t_i^{+}/t_i^{-})$ with the reference triple $(h_0,r,t_0)$ using our metric model, and the margin $\gamma$ is a hyperparameter to be tuned. For every training episode, we first sample one task/relation $T_r$ from the meta-training set $\mathbb{T}_{meta-training}$. Then from all the known triples in $T_r$, we sample one triple as the reference/training triple $D_r^{train}$ and a batch of other triples as the positive query/test triples $D_r^{test}$. The detail of the training process is shown in Algorithm~\ref{alg:training}. Our experiments are discussed in the next section.

\section{Experiments}
\subsection{Datasets}

\begin{table}[h!]
\small
\centering
 \begin{tabular}{c c c c c} \toprule
 Dataset & \# Ent. & \# R. & \# Triples & \# Tasks\\ 
 \midrule
NELL-One & 68,545 & 358 & 181,109 & 67 \\ 
Wiki-One & 4,838,244 & 822 &  5,859,240 & 183\\
 \bottomrule
 \end{tabular}
 \vspace{-1ex}
 \caption{Statistics of the Datasets. \# Ent. denotes the number of unique entities and \# R. denotes the number of all relations. \# Tasks denotes the number of relations we use as one-shot tasks.}
 \label{data_stats}
\end{table}

Existing benchmarks for knowledge graph completion, such as FB15k-237~\cite{toutanova2015representing} and YAGO3-10~\cite{mahdisoltani2013yago3} are all small subsets of real-world KGs. These datasets consider the same set of relations during training and testing and often include sufficient training triples for every relation. To construct datasets for one-shot learning, we go back to the original KGs and select those relations that do not have too many triples as one-shot task relations. We refer the rest of the relations as background relations, since their triples provide important background knowledge for us to match entity pairs.

Our first dataset is based on NELL~\cite{mitchell2018never}, a system that continuously collects structured knowledge by reading webs. We take the latest dump and remove those inverse relations. We select the relations with less than 500 but more than 50 triples\footnote{We want to have enough triples for evaluation.} as one-shot tasks. To show that our model is able to operate on large-scale KGs, we follow the similar process to build another larger dataset based on Wikidata~\cite{vrandevcic2014wikidata}. The dataset statistics are shown in Table~\ref{data_stats}. Note that the Wiki-One dataset is an order of magnitude larger than any other benchmark datasets in terms of the numbers of entities and triples. For NELL-One, we use 51/5/11 task relations for training/validation/testing. For Wiki-One, the division ratio is 133:16:34.

\subsection{Implementation Details}

\begin{table*}[t]
\small
\centering
 \begin{tabular}{lcccc|cccc} \toprule
& \multicolumn{4}{c}{\textbf{NELL-One}} & \multicolumn{4}{c}{\textbf{Wiki-One}}\\ 
\cmidrule{2-5} \cmidrule{6-9}
Model & MRR & Hits@10 & Hits@5 & Hits@1 & MRR & Hits@10 & Hits@5 & Hits@1 \\ \midrule
RESCAL & .071/.140 & .100/.229 & .082/.186 & \underline{.048/.089} & \underline{.119/.072} & \underline{.167/.082} & \underline{.132/.062} & \underline{.093/.051}\\
TransE & \underline{.082/.093} & \underline{.177/.192} & \underline{.126/.141} & .032/.043 & .023/.035 & .036/.052 & .029/.043 & .015/.025\\
DistMult & .075/.102 & .128/.177 & .093/.126 &  .045/.066 & .042/.048 & .086/.101 & .055/.070 & .017/.019\\
ComplEx & .072/.131 & .128/.223 & .041/.086 & .041/.086 & .079/.069 & {.148}/{.121} & .106/{.092} & .046/.040 \\
\midrule
GMatching (RESCAL) & {.144}/\textbf{.188} & .277/.305 & .216/.243 & {.087}/\textbf{.133} & .113/.139 & .330/.305 & .180/.228 & .033/.061\\
GMatching (TransE) & \underline{\textbf{.168}/.171} & .293/.255 & \underline{\textbf{.239}/.210} & \underline{\textbf{.103}/.122} & .167/.219 & .349/.328 & .289/.269 & .083/.163\\
GMatching (DistMult) & .119/.171 & .238/.301 & .183/.221 & .054/.114 & .190/\textbf{.222} & \underline{\textbf{.384}/\textbf{.340}} & \underline{\textbf{.291}/.271} & .114/\textbf{.164}\\
GMatching (ComplEx) & .132/.185 & \underline{\textbf{.308}/\textbf{.313}} & .232/\textbf{.260} & .049/.119 & \underline{\textbf{.201}/.200} & .350/.336 & .231/\textbf{.272} & \underline{\textbf{.141}/.120}\\
GMatching (Random) & .083/.151 & .211/.252 & .135/.186 & .024/.103 & .174/.198 & .309/.299 & .222/.260 & .121/.133\\
\bottomrule
 \end{tabular}
 \vspace{-1ex}
 \caption{Link prediction results on validation/test relations.
 KG embeddings baselines are shown at the top of the table and our one-shot learning (GMatching) results are shown at the bottom. 
 \textbf{Bold} numbers denote the best results on meta-validation/meta-test. \underline{Underline} numbers denote the model selection results from all KG embeddings baselines, or from all one-shot methods, i.e. selecting the method with the best validation score and reporting the corresponding test score.}
 \label{stats}
\end{table*}

In our experiments, we consider the following embedding-based methods: RESCAL~\cite{nickel2011three}, TransE~\cite{bordes2013translating}, DistMult~\cite{yang2014embedding} and ComplEx~\cite{trouillon2016complex}. For TransE, we use the code released by \newcite{lin2015learning}. For the other models, we have tried the code released by \newcite{trouillon2016complex} but it gives much worse results than TransE on our datasets. Thus we use our own implementations based on PyTorch~\cite{paszke2017automatic} for comparison. When evaluating existing embedding models, during training, we use not only the triples of background relations but also all the triples of the training relations and the one-shot training triple of those validation/test relations. However, since the proposed metric model does not require the embeddings of query relations, we only include the triples of the background relations for embedding training.
As TransE and DistMult use 1-D vectors to represent entities and relations, they can be directly used in our natching model. While for RESCAL, since it uses matrices to represent relations, we employ mean-pooling over these matrices to get 1-D embeddings. For the ComplEx model, we use the concatenation of the real part and imaginary part. The hyperparameters of our model are tuned on the validation task set and can be found in the appendix.

Apart from the above embedding models, a more recent method~\cite{dettmers2017convolutional} applies convolution to model relationships and achieves the best performance on several benchmarks. For every query $(h,r,?)$, their model enumerates the whole entity set to get positive and negative triples for training. We find that this training paradigm takes lots of computational resources when dealing with large entity sets and cannot scale to real-world KGs such as Wikidata\footnote{On a GPU card with 12GB memory, we fail to run their ConvE model on Wiki-One with batch size 1.} that have millions of entities. For the scalability concern, our experiments only consider models that use negative sampling for training.

\subsection{Results}
\begin{table*}[th]
\small
% \tiny
% \footnotesize
% \scriptsize
% \fontsize{7}{7.0}\selectfont
    \centering
    \begin{tabular}{lc|cc|cc}
    \toprule
    &  & \multicolumn{2}{c}{MRR} & \multicolumn{2}{c}{Hits@10} \\ 
\cmidrule{3-4} \cmidrule{5-6}
       Relations &  \# Candidates & GMatching & ComplEx & GMatching & ComplEx \\
    \midrule
       sportsGameSport & 123 & 0.424 & \textbf{0.466}($0.139^{\star}$) & \textbf{1.000} & 0.479($0.200^{\star}$)\\
       athleteInjuredHisBodypart & 299 & 0.025 & 0.026({$\mathbf{0.330^\star}$}) &  0.015 &  0.059($\mathbf{0.444^\star}$)\\
       animalSuchAsInvertebrate & 786 & 0.447 & 0.333($\mathbf{0.555^\star}$) & 0.626 &   0.587($\mathbf{0.783^\star}$)\\
       automobilemakerDealersInCountry & 1084 & \textbf{0.453} & 0.245($0.396^\star$) & \textbf{0.821} &  0.453($0.500^\star$)\\
       sportSchoolIncountry & 2100 & \textbf{0.534} & 0.324($0.294^\star$) & \textbf{0.745} &  0.529($0.571^\star$)\\
       politicianEndorsesPolitician & 2160 & 0.093 & 0.026($\mathbf{0.194^\star}$) & 0.226 &  0.047($\mathbf{0.357^\star}$) \\
       agriculturalProductFromCountry & 2222 & \textbf{0.120} & 0.029($0.042^\star$) & \textbf{0.288} &  0.058($0.086^\star$)\\
       producedBy & 3174 & 0.085 & 0.040($\mathbf{0.165^\star}$) & 0.179 &  0.075($\mathbf{0.241^\star}$)\\
       automobilemakerDealersInCity & 5716 & 0.026 &  0.024($\mathbf{0.041^\star}$) & 0.040  & 0.051($\mathbf{0.174^\star}$)\\
       teamCoach & 10569 & 0.017 & 0.065($\mathbf{0.376^\star}$) & 0.024 &  0.079($\mathbf{0.547^\star}$)\\
       geopoliticalLocationOfPerson & 11618 & 0.028 &  0.016($\mathbf{0.284^\star}$) & 0.035 &  0.035($\mathbf{0.447^\star}$)\\
    \bottomrule
    \end{tabular}
    \vspace{-0.02in}
    \caption{Results decomposed over different relations. ``$^\star$" denotes the results with standard training settings and ``\# Candidates" denotes the size of candidate entity set.}
    \label{tab:per_relation}
\end{table*}
The main results of our methods are shown in Table~\ref{stats}. We denote our method as ``GMatching'' since our model is trained to match local graph patterns. We use mean reciprocal rank (MRR) and Hits@K to evaluate different models. We can see that our method produces consistent improvements over various embedding models on these one-shot relations. The improvements are even more substantial on the larger Wiki-One dataset. To investigate the learning power of our model, we also try to train our metric model with randomly initialized embeddings. Surprisingly, although the results are worse than the metric models with pre-trained embeddings, they are still superior to the baseline embedding models. This suggests that, by incorporating the neighbor entities into our model, the embeddings of many relations and entities actually get updated in an effective way and provide useful information for our model to make predictions on test data.

It is worth noting that once trained, our model can be used to predict any newly added relations without fine-tuning, while existing models usually need to be re-trained to handle those newly added symbols. On a large real-world KG, this re-training process can be slow and highly computational expensive.

\paragraph{Remark on Model Selection} Given the existence of various KG embedding models, one interesting experiment is to incorporate model selection into hyper-parameter tuning and choose the best validation model for testing.

If we think about comparing KG embedding and metric learning as two approaches, the results from the model selection process can then be used as the ``final'' measurement for comparison.
For example, the baseline KG embedding achieves best MRR on Wiki-One with RESCAL (11.9\%), so we report the corresponding testing MRR (7.2\%) as the final model selection result for KG embedding approach.
In this way, at the top half of Table \ref{stats}, we select the best KG embedding method according to the validation performance. The results are highlighted with \underline{underlines}. Similarly, we select the best metric learning approach at the bottom.

Our metric-based method outperforms KG embedding by a large margin from this perspective as well. Taking MRR as an example, the selected metric model achieves 17.1\% on NELL-One and 20.0\% on Wiki-One; while the results of KG embedding are 9.3\% and 7.2\%. The improvement is 7.8\% and 12.8\% respectively.

% It is found that TransE does best on NELL-One on most of the measures, and RESCAL does best on Wiki-One on all measures.
% Similarly, we did the model selection on our one-shot methods at the bottom, where GMatching w/ TransE does best on NELL-One, and the methods with DistMult and ComplEx work best for different measures on Wiki-One.

\subsection{Analysis on Neighbor-Encoder}
\begin{figure}[t]
\centering
\includegraphics[width=1.0\linewidth]{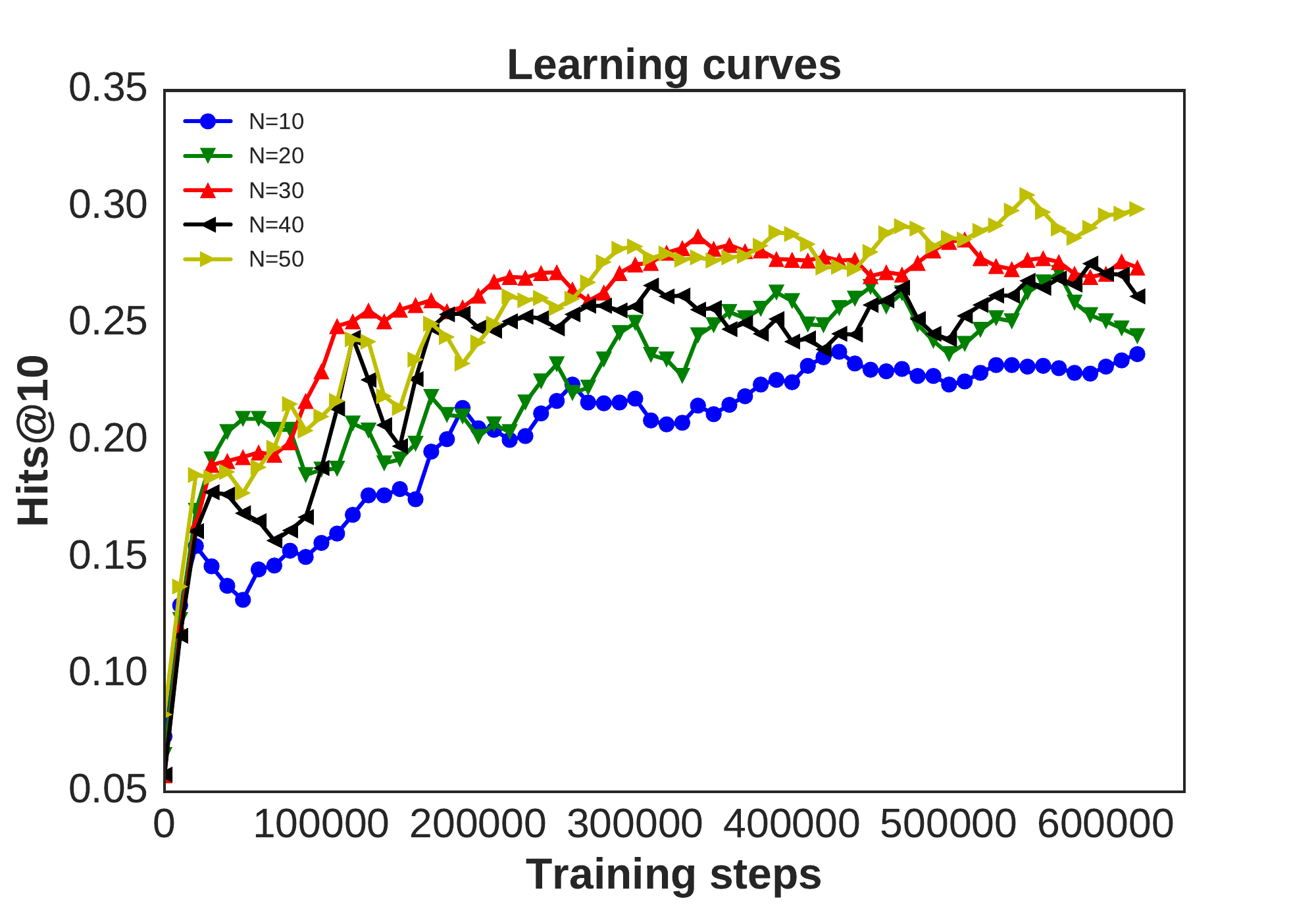}
 \vspace{-0.3in}
\caption{The learning curves on NELL-one. Every run uses different number of neighbors. The y-axis is Hits@10 calculated on all the validation relations. }
\label{curves}
\end{figure}

As our model leverages entities' local graph structures by encoding the neighbors, here we try to investigate the effect of the neighbor set by restricting the maximum number of neighbors. If the size of the true neighbor set is larger than the maximum limit, the neighbors are then selected by random sampling. Figure~\ref{curves} shows the learning curves of different settings. These curves are based on the Hits@10 calculated on the validation set. We see that encoding more neighbors for every entity generally leads to better performance. We also observe that the model that encodes 40 neighbors in maximum actually yields worse performance than the model that only encodes 30 neighbors. We think the potential reason is that for some entity pairs, there are some local connections that are irrelevant and provide noisy information to the model.

\subsection{Ablation Studies}
\begin{table}[t]
\small
    \centering
    \begin{tabular}{l|c}
    \toprule
       Configuration & Hits@10\\
    \midrule
       Full Model with ComplEx & \textbf{.308}/\textbf{.313}\\
    \midrule 
       w/o Matching Processor & .266/.269 \\
       w/o Neighbor Encoder & .248/.296\\
       w/o Scaling Factor & .229/.219 \\
    \bottomrule
    \end{tabular}
    \vspace{-1ex}
    \caption{Ablation on different components.}
    \label{tab:ablation}
\end{table}

We conduct ablation studies using the model that achieves the best Hits@10 on the NELL-One dataset. The results are shown in Table~\ref{tab:ablation}. We use Hits@10 on validation and test set for comparison, as the hyperparameters are selected using this evaluation metric. We can see that both the matching processor\footnote{Matching without Matching Processor is equivalent to matching using simple cosine similarity.} and the neighbor encoder play important roles in our model. Another important observation is that the scaling factor $1/\mathcal{N}_e$ turns out to be very essential for the neighbor encoder. Without scaling, the neighbor encoder actually gives worse results compared to the simple embedding-based matching.

\subsection{Performance on Different Relations}
When testing various models, we observe that the results on different relations are actually of high variance. Table~\ref{tab:per_relation} shows the decomposed results on NELL-One generated by our best metric model (GMatching-ComplEx) and its corresponding embedding method. For reference, we also report the embedding model's performance under standard training settings where $75\%$ of the triples (instead of only one) are used for training and the rest are used for testing. We can see that relations with smaller candidate sets are generally easier and our model could even perform better than the embedding model trained under standard settings. For some relations such as \textit{athleteInjuredHisBodypart}, their involved entities have very few connections in KG. It is as expected that one-shot learning on these kinds of relations is quite challenging. Those relations with lots of ($>$3000) candidates are challenging for all models. Even for embedding model with more training triples, the performance on some relations is still very limited. This suggests that the knowledge graph completion task is still far from being solved.

\section{Conclusion}
This paper introduces a one-shot relational learning framework that could be used to predict new facts of long-tail relations in KGs. Our model leverages the local graph structure of entities and learns a differentiable metric to match entity pairs. In contrast to existing methods that usually need finetuning to adapt to new relations, our trained model can be directly used to predict any unseen relation and also achieves much better performance in the one-shot setting. Our future work might consider incorporating external text data and also enhancing our model to make better use of multiple training examples in the few-shot learning case.

\section*{Acknowledgments}
This research is supported by an IBM Faculty Award. We also thank the anonymous reviewers for their useful feedback.

\bibliography{emnlp2018}
\bibliographystyle{acl_natbib_nourl}

\clearpage
\appendix

\section{Hyperparameters}
 For the NELL dataset, we set embedding size as 100. For Wikidata, we set the embedding size as 50 for faster training with millions of triples. The embeddings are trained for 1,000 epochs. The other hyperparamters are tuned using the Hits@10 metric\footnote{The percentage of correct answer ranks within top10.} on the validation tasks. For matching steps, the optimal setting is 2 for NELL-One and 4 for Wiki-One. For the number of neighbors, we find that the maximum limit 50 works the best for both datasets. For parameter updates, we use Adam~\cite{kingma2014adam} with the initial learning rate 0.001 and we half the learning rate after 200k update steps. The margin used in our loss function is 5.0. The dimension of LSTM's hidden size is 200.
 
 \section{Few-Shot Experiments}

\begin{table}[h]
\small
    \centering
    \begin{tabular}{l|cc|cc}
    \toprule
      Metrics & \multicolumn{2}{c|}{GMatching}  & \multicolumn{2}{c}{ComplEx} \\
      & 1-shot &5-shot &1-shot &5-shot \\
    \midrule
      MRR & .132/.185 &{.178}/{.201} & .072/.131 & .113/.200\\
      Hits@10 & .308/.313 & {.307}/.311 & .128/.223 & .221/{.325}\\
      Hits@5 &.232/.260 & {.241}/.264 & .041/.086 & .160/{.269}\\
      Hits@1 &.049/.119 & .109/{.143} & .041/.086 & .113/{.133}\\
    \bottomrule
    \end{tabular}
    \caption{5-shot experiments on NELL-One.}
    \label{tab:5-shot}
\end{table}

% % %=====================Original Version========================
% % % \begin{table}[t]
% % % \small
% % %     \centering
% % %     \begin{tabular}{l|cccc}
% % %     \toprule
% % %       Metrics & \multicolumn{2}{|c|}{GMatching}  & \multicolumn{2}{|c|}{ComplEx} \\
% % %     \midrule
% % %       MRR & &\underline{.178}/\underline{.201} & & .113/.200\\
% % %       Hits@10 & \underline{.307}/.311 & .221/\underline{.325}\\
% % %       Hits@5 & \underline{.241}/.264 & .160/\underline{.269}\\
% % %       Hits@1 & .109/\underline{.143} & .113/\underline{.133}\\
% % %     \bottomrule
% % %     \end{tabular}
% % %     \caption{5-shot experiments on NELL-One.}
% % %     \label{tab:5-shot}
% % % \end{table}

Although our model is designed for one-shot settings, it can be directly applied to the $k$-shot ($k>1$) setting by applying an aggregation function over the hidden states or scores of all the $k$ examples in the support set. 

This section considers the 5-shot case on NELL-One, and simply ranks the candidates with the maximum of the five scores $score^{s_i}_{K}$ according to Eq. \ref{eq:scoring}, where $s_i$ ($i \in {1,5}$) are 5 support entity-pairs. Table~\ref{tab:5-shot} shows the results. We can see that with more training triples, the performance of the KG embedding model improves a lot, while our model's improvement is limited. We think this is because (1) our model handles each example individually and ignores their interactions; (2) our model does not perform any training on the meta-testing relations, while the KG embedding methods benefit more from re-training when the number of labeled tuples $k$ increases. Future research could consider using more advanced aggregation module such as attention to model the interaction of multiple training examples; as well as exploring meta-learning approaches \cite{ravi2017optimization} to enable fast adaptation of metrics during meta-testing.

\end{document}